\documentclass[letterpaper, 10 pt, conference]{ieeeconf}  

\IEEEoverridecommandlockouts                              

\usepackage{graphics} 
\usepackage{epsfig} 
\usepackage{amsmath} 
\usepackage{amssymb}  
\usepackage{booktabs}
\usepackage[dvipsnames]{xcolor}
\usepackage{url}
\usepackage{url}
\usepackage{cite}

\title{\LARGE \textbf{Multi-segment Soft Robot Control via \\ Deep Koopman-based Model Predictive Control}
}

\author{Lei Lv$^{1,2,*}$, Lei Liu$^{2,3,*}$, Lei Bao$^{3,4,*}$, Fuchun Sun$^{2,\dag}$,~\IEEEmembership{Fellow,~IEEE}, Jiahong Dong$^{5,\dag}$, \\Jianwei Zhang$^{6}$,~\IEEEmembership{Member,~IEEE},  Xuemei Shan$^4$, Kai Sun$^{2,3}$, Hao Huang$^{7}$, Yu Luo$^{2,8}$
\thanks{*These authors contributed equally to this research.}
\thanks{$^{\dag}$Co-corresponding authors: Fuchun Sun (\protect\url{fcsun@tsinghua.edu.cn}) and Jiahong Dong (\protect\url{dongjiahong@mail.tsinghua.edu.cn}).}
\thanks{$^{1}$Shanghai Research Institute for Intelligent Autonomous Systems, Tongji University. $^{2}$Department of Computer Science and Technology, Tsinghua  University. $^{3}$School of Biomedical Engineering, Tsinghua University. $^{4}$Beijing Soft Robot Tech Co., Ltd. $^5$School of Clinical Medicine, Tsinghua University. $^6$Department of Informatics, University of Hamburg. $^7$School of Mechanical Engineering and Automation, Beihang University.$^8$Huawei Noah's Ark Lab.}
}

\begin{document}

\maketitle
\thispagestyle{empty}
\pagestyle{empty}

\begin{abstract}
Soft robots, compared to regular rigid robots, as their multiple segments with soft materials bring flexibility and compliance, have the advantages of safe interaction and dexterous operation in the environment. However, due to its characteristics of high dimensional, nonlinearity, time-varying nature, and infinite degree of freedom, it has been challenges in achieving precise and dynamic control such as trajectory tracking and position reaching. To address these challenges, we propose a framework of Deep Koopman-based Model Predictive Control (DK-MPC) for handling multi-segment soft robots. We first employ a deep learning approach with sampling data to approximate the Koopman operator, which therefore linearizes the high-dimensional nonlinear dynamics of the soft robots into a finite-dimensional linear representation. Secondly, this linearized model is utilized within a model predictive control framework to compute optimal control inputs that minimize the tracking error between the desired and actual state trajectories. The real-world experiments on the soft robot “Chordata” demonstrate that DK-MPC could achieve high-precision control, showing the potential of DK-MPC for future applications to soft robots. More visualization results can be found at \textcolor{Cerulean}{\protect\url{https://pinkmoon-io.github.io/DKMPC/}}.
\end{abstract}

\section{INTRODUCTION}



Soft robots, characterized by their multiple segments connected and actuated by soft materials, offer unparalleled flexibility and compliance, enabling them to interact safely and adaptively with humans and their environments~\cite{rus2015design,laschi2022embodied}. This unique capability has positioned soft robots as promising candidates in a variety of applications, including rehabilitation~\cite{tang2022learning}, minimally invasive surgeries~\cite{kwok2022soft}, and underwater operations~\cite{xie2023octopus}. Soft robots are also ideal platforms for embodied intelligence~\cite{liu2025embodied}. Despite their potential, the development of soft robots is challenged by the inherent nonlinearity of soft materials, the complex coupling between segments, and the time-varying dynamics that come with infinite degrees of freedom. These characteristics make it exceedingly difficult to precisely model the robots' complex dynamics and design controllers by traditional methods~\cite{rus2015design,george2018control,wang2022control}.


To address these challenges,  two primary control strategies have been widely adopted: model-based control and learning-based control~\cite{haggerty2023control}. Model-based control approaches often utilize pseudo-rigid models~\cite{roesthuis2016steering}, constant curvature models~\cite{jones2006kinematics}, or material property-based models~\cite{gao2016mechanical} to simplify the characteristics of soft robots under various assumptions. These methods reconstruct dynamic models specific to different robots, offering a structured way to predict and control their behavior. However, there is often a significant gap between these simplified models and the actual robots, which can result in suboptimal performance, particularly when it comes to achieving highly precise control~\cite{padmanabhan2015closed,wang2022data}.




\begin{figure}[!t]
\centering
\includegraphics[width=0.48\textwidth]{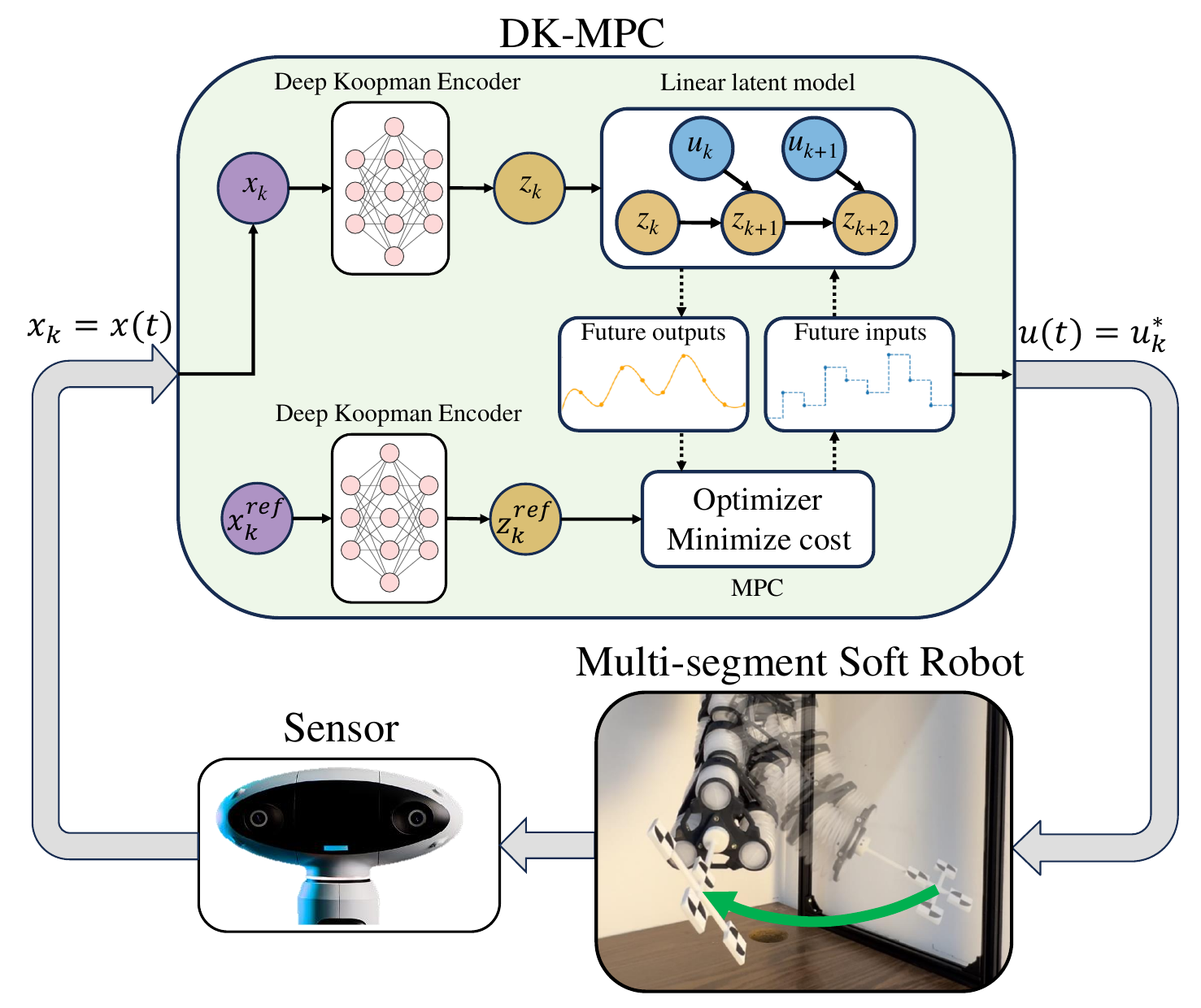}
\caption{Illustration of the proposed Deep Koopman-based Model Predictive Control (DK-MPC) framework for a multi-segment soft robot. The deep Koopman operator maps both the reference state \( x^{{ref}} \) and the state \( x \) into a high-dimensional linear latent space. Based on the latent states and the linear dynamics, an MPC controller is employed to generate the optimal control signals \( u^* \), ensuring the end-effector of the soft robot follows the reference trajectory.}
\label{fig_1}
\end{figure}

The limitations of these traditional model-based control methods have led to the exploration of learning-based control strategies, which leverage deep learning methods to capture the nuanced dynamics or kinematics of soft robots more effectively~\cite{laschi2023learning}. One approach in learning-based control involves the use of inverse  models~\cite{george2018control,li2024disturbance,george2017learning,wiese2021transfer}, which directly map operational space to actuation space, 
simplifying the design of controllers.
However, the inherent variability in soft robot behavior and the multiple solutions to the inverse dynamics or kinematics problem can still pose difficulties for learning algorithms. On another front, reinforcement learning (RL) has been explored for soft robot control, with policies trained in simulated environments~\cite{coevoet2017software} or through the use of learned forward models~\cite{zhou2024cable}. While RL holds the potential to handle the stochastic nature of soft robots and discover effective control policies through interaction with the environment, it often requires a substantial amount of data and experiences the sim-to-real transfer gap, where policies that perform well in simulation can not guarantee real-world performance. 

To address the complexities of forward modeling in soft robots, the Koopman operator-based method has been effectively utilized, offering a novel perspective for control design~\cite{bruder2021koopman,bruder2019nonlinear,wang2022improved,bruder2020data}. The Koopman operator enables the transformation of a nonlinear system into a linear representation, which is significantly beneficial for nonlinear soft robot control. However, a critical challenge lies in the selection of appropriate lifting functions\cite{shi2023koopman}, as inadequate choices can lead to serious modeling errors, impacting the model's predictive accuracy and the controller's performance~\cite{chen2023control}.



In this paper, we propose a deep Koopman-based model predictive control (DK-MPC) framework for multi-segment soft robots, whereas previous similar frameworks have primarily focused on rigid robots and were limited to simulation implementations. Our approach automates the learning of suitable embeddings, thereby enhancing the Koopman operator's predictive quality. By integrating deep neural networks (DNNs) into the operator, we achieve a globally linearized model of the robot, which provides the system dynamic constraint for precise position control through Model Predictive Control (MPC), as shown in Fig.~\ref{fig_1}. This integration allows for real-time optimization of control inputs, ensuring precise control performance for soft robots.

Real-world experiments conducted on the soft robot ``Chordata'' demonstrate the efficacy of the proposed DK-MPC method. The system achieves highly precise control, showcasing a significant improvement over traditional control methods. This advancement indicates the potential of DK-MPC for future soft robot applications, particularly in tasks requiring high accuracy and adaptability~\cite{wang2022improved}. And, as a data-driven method, DK-MPC can be seamlessly tailored into different multi-segment soft robots, providing a promising solution for soft robot applications.

\begin{figure*}[!t]
\centering
\includegraphics[width=0.9\textwidth]{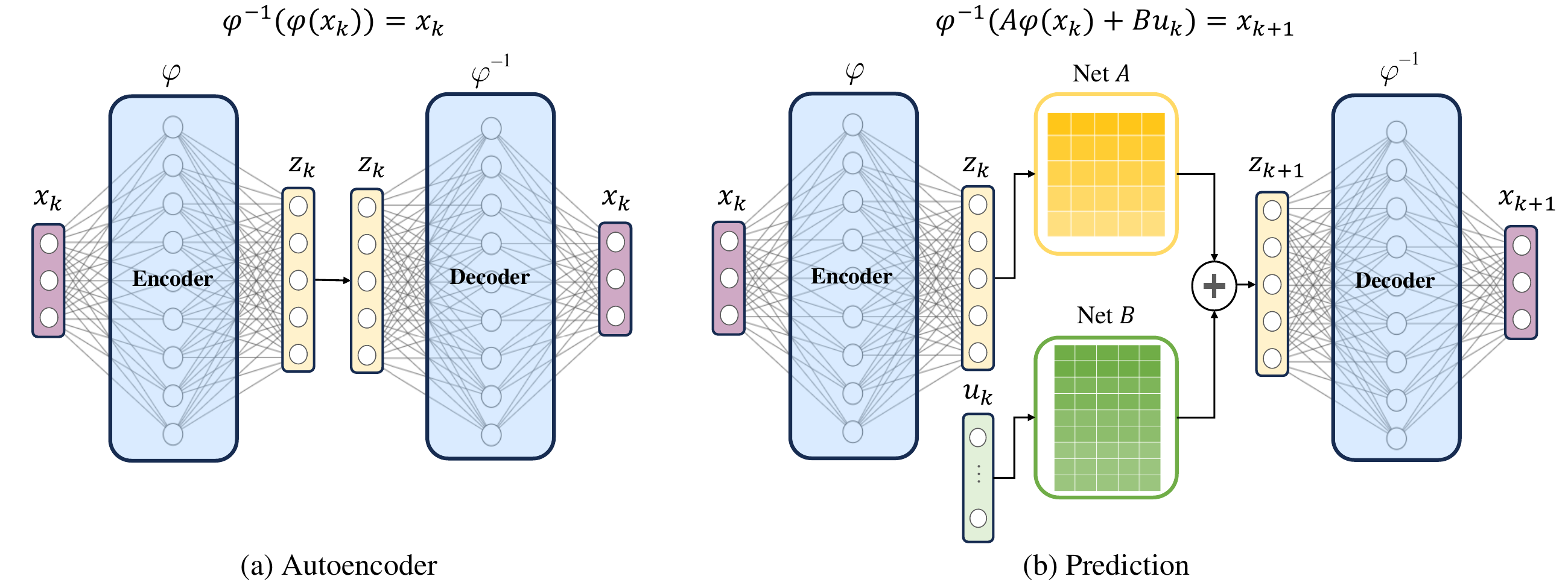}
\caption{Design of deep Koopman operator architecture in Sec.~2.C. 
(a) Deep auto-encoder for learning lifting function $\varphi$ and its inverse function $\varphi^{-1}$. 
(b) Illustration of the learning process for linear operator $A$ and control-affine matrix $B$.}
\label{fig_4}
\end{figure*}

\section{Deep Koopman-based Model Predictive Control}
In this section, we delve into the detailed methodology of the proposed DK-MPC, which is designed to address the control challenges of multi-segment soft robots. We first introduce the overall framework, then illustrate the Koopman operator's basis and learning method in the high-dimensional situation, and finally give a control strategy by incorporating the MPC with the deep Koopman operator.

\subsection{Overall Framework of DK-MPC}

The framework of the proposed DK-MPC is illustrated in Fig. 1. A deep learning-based Koopman operator is employed to construct a global linear time-invariant embedding of the dynamic system by elevating the original state space to a higher-dimensional space. At each control step, the deep Koopman encoder transforms the reference state $x^{ref}$ into a high-dimensional control reference. And, the actual state $x$ is measured by a visual sensor and then elevated to a high-dimensional space using the same encoder.  Within this space, an MPC controller minimises the quadratic cost of the latent states $z$ and $z^{ref}$ over a finite time horizon to generate the optimal control signals $u^*$ to track the reference. 

\subsection{Koopman Operator for Linearizing Soft Robot Dynamics}

Consider a discrete-time nonlinear system characterized by the function $f$,  which evolves the state of the multi-segment soft robot. Here, $x_k$ represents the state of the system at time step $k$ and $u_k$ is the control input.
\begin{equation}
	\label{1}
	{
        {x}_{k + 1} = {f}({x}_k, {u}_k), 
    }
\end{equation}

Given the complexity and nonlinearity of $f$, direct control design is often intractable. To circumvent this, we employ the Koopman operator, which acts on a lifted space defined by a function $\varphi$. This lifting function $\varphi$ maps the original nonlinear dynamics into a higher-dimensional feature space where the dynamics are linearizable:
\begin{equation}
\label{2}
\varphi\left( {f}\left( {x}_k, {u}_k \right), {u}_{k+1} \right) = \mathcal{K} \varphi\left( {x}_k, {u}_k \right),
\end{equation}

The choice of $\varphi$ is critical since it determines the effectiveness of the linear representation and the subsequent control design. In this work, we utilize a deep learning-based approach to approximate $\varphi$, allowing us to capture the complex nonlinear dynamics inherent in the soft robots.




Building upon the Koopman operator framework, we further decompose the lifting function $\varphi$ into state-dependent \( \varphi_{x} \) and input-dependent parts \( \varphi_{u} \),

\begin{equation}
	\label{3}
	{
        \varphi({x},{u}) = [\varphi_{x}({x});\varphi_{u}({u})],
    }
\end{equation}
which allows us to rewrite the dynamics in a control-affine form with matrix multiplication
\begin{equation}
	\label{4}
	{
        \left[ \begin{array}{c}\varphi_{{x}}{(x_{k+1})}\\
 \varphi_{u}({u_{k+1}}) \\
 \end{array}\right] =\left[ \begin{array}{cc}K_{xx} & K_{xu}\\
 K_{ux} & K_{uu} \\
 \end{array} \right] \left[ \begin{array}{c}\varphi_{x}{(x_{k})}\\
 \varphi_{u}({u_{k}}) \\
 \end{array} \right],
    }
\end{equation}
Thus, we can obtain
\begin{equation}
	\label{5}
	{
        \varphi_{x}({x_{k+1}}) = K_{xx}\varphi_{x}{(x_{k})}+K_{xu}\varphi_{u}{(u_{k})},
    }
\end{equation}

Following the simplifications and notations introduced in previous works \cite{bruder2020data,shi2022deep}, the control component of the lifting function can be represented directly by the control input itself, i.e., $\varphi_{u}({u}) = {u}$.  We then denote the linear operators as $K_{xx} = A$ and $K_{xu} = B$, and the lifted state as $\varphi_{x}(x_{k}) = z_{k}$. This leads us to the globally linearized representation of the original nonlinear dynamics
\begin{equation}
	\label{6}
	{        z_{k+1} = Az_{k}+Bu_{k}.     }
\end{equation}

With this formulation, we have successfully transformed the original nonlinear dynamics (Eq.~\ref{1}) into a linear state-space representation (Eq.~\ref{6}). This linearized model is not only more amenable to control design but also captures the global behavior of the original nonlinear multi-segment soft robot system. 


\subsection{Deep Learning Method for Koopman Operator}

In this subsection, we detail the deep learning method to approximate the proposed Koopman operator and the associated lifting functions. As illustrated in Fig. 2, our approach involves the construction of a deep auto-encoder framework to learn the lifting function \(\varphi\) and its corresponding inverse function \(\varphi^{-1}\). The auto-encoder consists of two main components: an encoder and a decoder. The encoder \(\varphi\), implemented as a Multi-Layer Perceptron (MLP), transforms the original state space into a higher-dimensional latent space where the dynamics can be approximated linearly. The decoder \(\varphi^{-1}\), on the other hand, reconstructs the original states from this latent space.

Additionally, we utilize two single-layer MLPs \textbf{\textit{without biases and activation function}} to approximate the linear operator \(A\) and the control-affine matrix \(B\) in Eq.~(\ref{6}), where the absence of biases and activation in these networks allows for a simple matrix representation. These components are essential for capturing the linear dynamics within the latent space, aligning with the theoretical framework of the Koopman operator.

                                                  

\textbf{Reconstruction Loss:} To ensure the auto-encoder precisely captures the relationship between the original and latent spaces, we define a reconstruction loss with $L_2$ norm 
\begin{equation}
	\label{7}
	{
       L_{\text{recon}}= \left\| {{{x_k}} - \varphi ^{ - 1}(\varphi ({{x_k}}))} \right\|^2_2,
    }
\end{equation}
\textbf{Linear Dynamics Loss:} To learn the linear operator \(A\) and the control-affine matrix \(B\), we introduce a loss function that enforces the latent dynamics to adhere to a linear model
\begin{equation}
	\label{8}
	{
       L_{\text{linear}} = \left\| \varphi ({x}_{k + 1}) - (A \varphi({x}_{k}) + B {u_{k}}) \right\|^2_2,  
    }
\end{equation}
\textbf{Prediction Loss:} To enhance the model's predictive capability over longer time horizons and enable precise future state predictions, we define a prediction loss over $m$ steps state prediction by the linear dynamics

\begin{align}
L_{\text{pred}} &= \left\| \varphi({x}_{k+m}) - {z}_{k+m} \right\|^2_2, \ \text{where} \nonumber \\
{z}_{k+m} &= A^m {z}_k + A^{m-1}B {u}_{k} + \cdots + B {u}_{k+m-1}. 
\end{align}

This formulation allows the model to capture the sequential application of the linear operator \(A\) and control inputs \(B\), providing a robust mechanism for multi-step future predictions.

Our final loss function, as defined in Eq. (10), is a weighted sum of the reconstruction loss, prediction loss, and linear loss, complemented by an $L_2$ regularization term to prevent overfitting~\cite{ng2004feature},
\begin{equation}
	\label{10}
	{
       L = \lambda_{1} L_{recon} + \lambda_{2}  L_{pred} + \lambda_{3} L_{linear} + \lambda_{4} \Vert W \Vert_2^2.
    }
\end{equation}

This comprehensive framework provides an end-to-end training method for simultaneously training the linear operator \( A\), the control affine matrix \( B \), and the lifting function \(\varphi\), thereby enabling the effective control of the nonlinear dynamics of multi-segment soft robots.
\subsection{Integrating MPC with Deep Koopman Operator}

With the learnable Koopman Operator, we integrate Model Predictive Control (MPC) to control the nonlinear dynamics of multi-segment soft robots. MPC is renowned for its ability to handle systems with constraints and its capacity to optimize control actions based on predicting future behavior~\cite {10610952,luo2024robust}. By leveraging the linearized model derived from the Koopman operator, we can implement an MPC strategy that is both effective and computationally efficient.


\begin{figure*}[!t]
\centering
\includegraphics[width=7.0in]{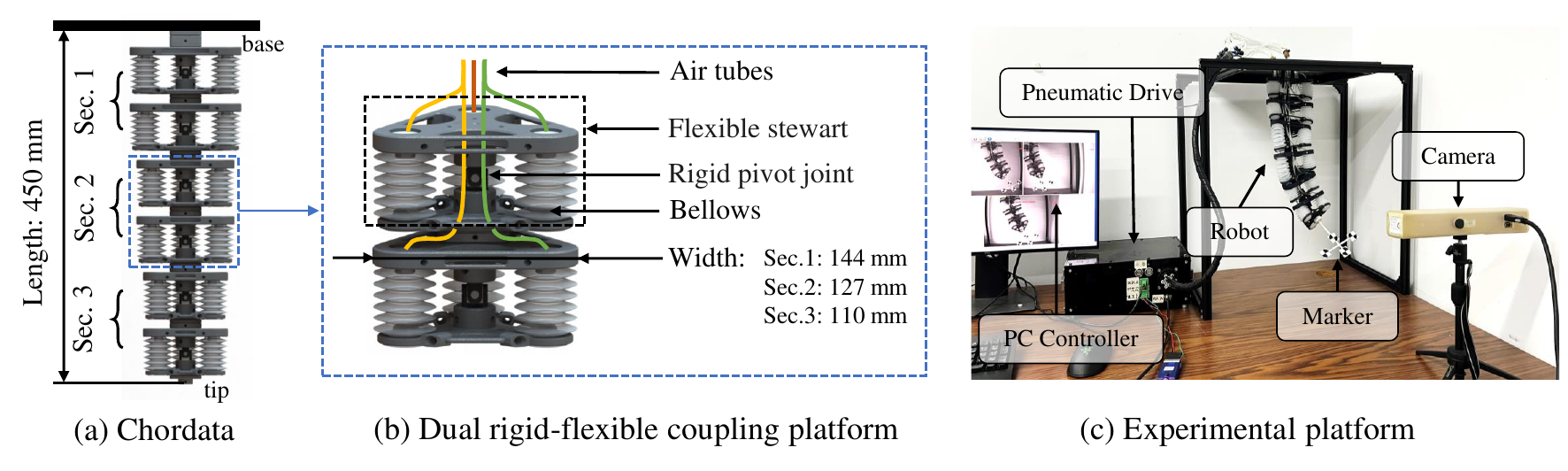}
\caption{Design of the multi-segment soft robot “Chordata”. (a) Overview of the robot, which has $3$ independent segments and a total length of $450 mm$. The addition of bones along the joints enhances stability and stiffness. (b) Detailed segment design: a flexible Stewart platform with three bellows in a circular array, constrained by a central rigid pivot. (c) Experimental platform setup, including a stereo camera (\textit{MicronTracker H3-60}) for tip position tracking, a pneumatic drive controlling air pressure, and PC controller for data processing, system identification and control.}
\label{fig_2}
\end{figure*}

The optimization problem within the MPC framework is to find the sequence of control inputs $u^*_{t: t+H}$ that minimize the cost function. This function is a sum of weighted state deviations from the desired state and the control efforts with the linear dynamics model constraint, formulated as


\begin{equation}  
    \label{11}
    \begin{aligned}  
    & \min_{\hat{u}_{t:t+H}} \sum_{k=0}^{H} \left(z_{t+k} - z_{t+k}^{ref}\right)^T \hat{Q} \left(z_{t+k} - z_{t+k}^{ref}\right) + \hat{u}_{t+k}^T \hat{R} \hat{u}_{t+k} \\  
    & \text{s.t.} \quad  
    \begin{cases}  
    z_{t+k} = A z_{t+k-1} + B \hat{u}_{t+k-1}, & k = 1, 2, \ldots, H \\  
    z_t = \phi(x_t), \\
    \ z^{ref}_{t+k} = \phi\left(x^{ref}_{t+k}\right), &
    k = 1, 2, \ldots, H\\  
    u_{\min} \leq \hat{u}_{t+k} \leq u_{\max}, & k = 0, 1, 2, \ldots, H  
    \end{cases}  
    \end{aligned}  
\end{equation}
where \( H \in \mathbb{N} \) is the prediction horizon, indicating the number of prediction steps over which the MPC controller plans. The matrix \( \hat{Q} \) is an \( n \times n \) positive semi-definite matrix used to penalize the deviation of the state \( z_t \) from the desired state \( z_t^{ref} \). Here, $n$ represents the dimension of the latent variable. \( \hat{R} \) is an \( m \times m \) positive semi-definite matrix used to penalize the control input \( \hat{u}_t \). The symbols \( {u}_{\min} \) and \( {u}_{\max} \) represent the minimum and maximum allowable values for the control input, respectively. 

At each control step, we first sample the real states $x_t$ and the reference state $x^{ref}_{t:t+H}$ of the nonlinear system and encode it to the high-dimensional latent state $z_t$ and $z^{ref}_{t:t+H}$, then solve the optimization problem~(\ref{11}) to obtain the optimal control input sequence $u^*_{t:t+H}$. Only the first element $u^*_t$ would be applied to the original system and we repeat this process until the system converges.

\section{Experiment}
In this section, we describe the process of soft robot system design, data collection, and model training, and implement the DK-MPC controller to control the soft robot ‘Chordata’. Afterward, we demonstrate the feasibility of DK-MPC through two experiments: a path-tracking task and a moving-target tracking task, where the former demonstrates the accuracy and high dynamic response of DK-MPC, and the latter shows that DK-MPC is capable of dynamic control, and is expected to be applied in the future.

\begin{table}[!t]
    \caption{Hyperparameters for the Proposed Deep Learning Model}
    \label{tab:hyperparameters}
    \centering
    \begin{tabular}{lc}
        \toprule
        Hyperparameter & Value \\
        \midrule
        Learning Rate ($\eta$) & $0.001$ \\

        Batch Size ($B$) & $64$ \\

        Latent Space Dimension ($n$) & $12$ \\

        Number of Hidden Layers (Encoder) & $2$ \\

        Number of Neurons (Encoder) & \{3, 128, 256, 12\} \\

        Activation Function (Encoder) & ReLU \\

        Number of Hidden Layers (Decoder) & $2$ \\

        Number of Neurons (Decoder) & \{12, 128, 256, 3\} \\

        Activation Function (Decoder) & ReLU \\

        Number of Neurons (Net A) & \{12, 12\} \\

        Number of Neurons (Net B) & \{9, 12\} \\

        Optimizer & Adam \\
        \bottomrule
    \end{tabular}
\end{table}

\subsection{Design of the Multi-segment Soft Robot}


To empirically validate the modeling and control theories presented earlier, we have engineered a multi-segment rigid-flexible coupling soft robot named ‘Chordata’. As depicted in Fig.~\ref{3}, the robot spans a total length of $450 mm$ and is segmented into three independently actuated sections. Each section is equipped with a dual rigid-flexible coupling platform, harnessing pneumatic pressure to regulate the inflation and deflation of three circumferentially arranged sets of bellows, thereby achieving rotational control of the segment.

To improve the robot's stability and rigidity, we have integrated spinal elements along the central axis of each segment, culminating in a structure that is both rigid and flexible. The extremity of the robot is adorned with a flange, intended for the attachment of an end-effector. The design features a tapered width, decreasing from the base to the tip, to enhance the load-bearing capacity at the base while preserving flexibility at the tip.


The pneumatic drive controls the robot through nine independent air pressure channels managed by proportional valves, with pressure ranging from $0kPa$ to $40kPa$ for bellows actuation.  A PC controller can orchestrate the pressure commands and manage system identification, control algorithms, and data processing. A stereo camera system, \textit{MicronTracker H3-60}, is employed to track the position of the robot's tip by monitoring markers attached to it. This positional feedback is relayed to the PC controller, enabling real-time adjustments and ensuring the end-effector's trajectory adheres to the desired path.
\begin{figure}[!t]
\centering
\includegraphics[width=0.5\textwidth]{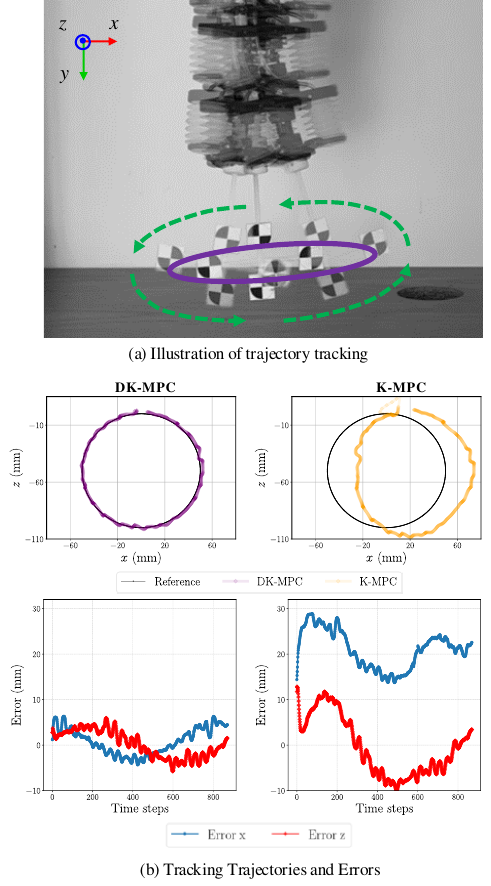}
\caption{Visualization of trajectory tracking. (a) Task execution of the soft robot, where the green dashed arrow is the direction and the elliptical arc is the trajectory. (b) The results and errors of the proposed DK-MPC and  K-MPC controllers during circular trajectory tracking. }
\label{fig_7}
\end{figure}

\subsection{Data Collection}

We adopt a randomized actuation strategy for the pneumatic actuators, ensuring a wide-ranging and diverse dataset. The stereo camera system continuously monitors the marker attached to the tip of the robot, providing high-fidelity positional information in three-dimensional space. Concurrently, we recorded the corresponding pneumatic actuation pressures, compiling a dataset of 45,607 tuples formatted as  \((x_k, u_k, x_{k+1})\), where \(x_k \in \mathbb{R}^3\) represents the current position in three-dimensional space captured by the visual sensor, \(u_k \in \mathbb{R}^9\) represents the actuation pressures in nine pneumatic chambers at time step \(k\), and \(x_{k+1}\) is the resulting state after applying the actuation pressure.



\subsection{Model training}
The dataset was partitioned into training, testing, and validation sets. All data were normalized by the Min-Max normalization technique~\cite{goodfellow2016deep}, where each feature's range is scaled to \([-1, 1]\) by following equation,
\begin{equation}
	\label{12}
	x'_i = 2 \left( \frac{x_i - \min(x)}{\max(x) - \min(x)} \right) - 1
\end{equation}
This normalization can ensure the model training process remains stable and consistent~\cite{goodfellow2016deep}.

The hyperparameters used for the proposed architecture are summarized in Table~\ref{tab:hyperparameters}, where these settings were chosen to optimize the performance and stability of the model during the training process.

\begin{table}[!t]
    \caption{Error Comparison of Methods for Trajectory-Tracking Tasks (in Millimeters)}
    \label{tab:methods_comparison}
    \centering
    \begin{tabular}{lcccccc}
        \toprule
        \textbf{Controller} & \textbf{``O''} & \textbf{``T''} & \textbf{``H''} & \textbf{``U''} & \textbf{Avg. Err. } \\
        \midrule 
        DK-MPC     & 2.79   & 3.38   & 3.15   & 3.13   & 3.11    \\
        K-MPC      & 22.03   & 18.50   & 24.04   & 24.97   & 22.49     \\
        \bottomrule
    \end{tabular}
\end{table}

\subsection{Tracking of Trajectory}

\begin{figure*}[!t]
    \centering
    \includegraphics[width=0.75\textwidth]{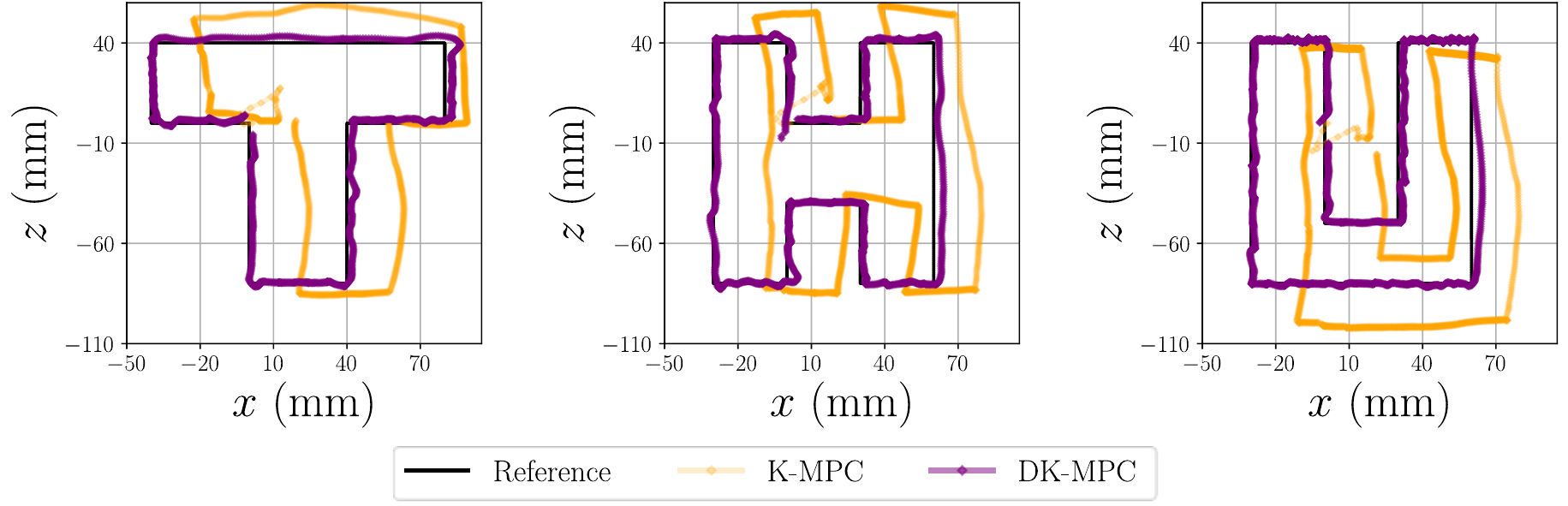}
    \caption{The results of DK-MPC and K-MPC in 'T', 'H', and 'U' trajectory tracking tasks. The reference trajectories are displayed in purple, and the actual trajectories are shown in orange.}
    \label{fig_8}
\end{figure*}
\begin{figure*}[!t]
    \centering
    \includegraphics[width=0.9\textwidth]{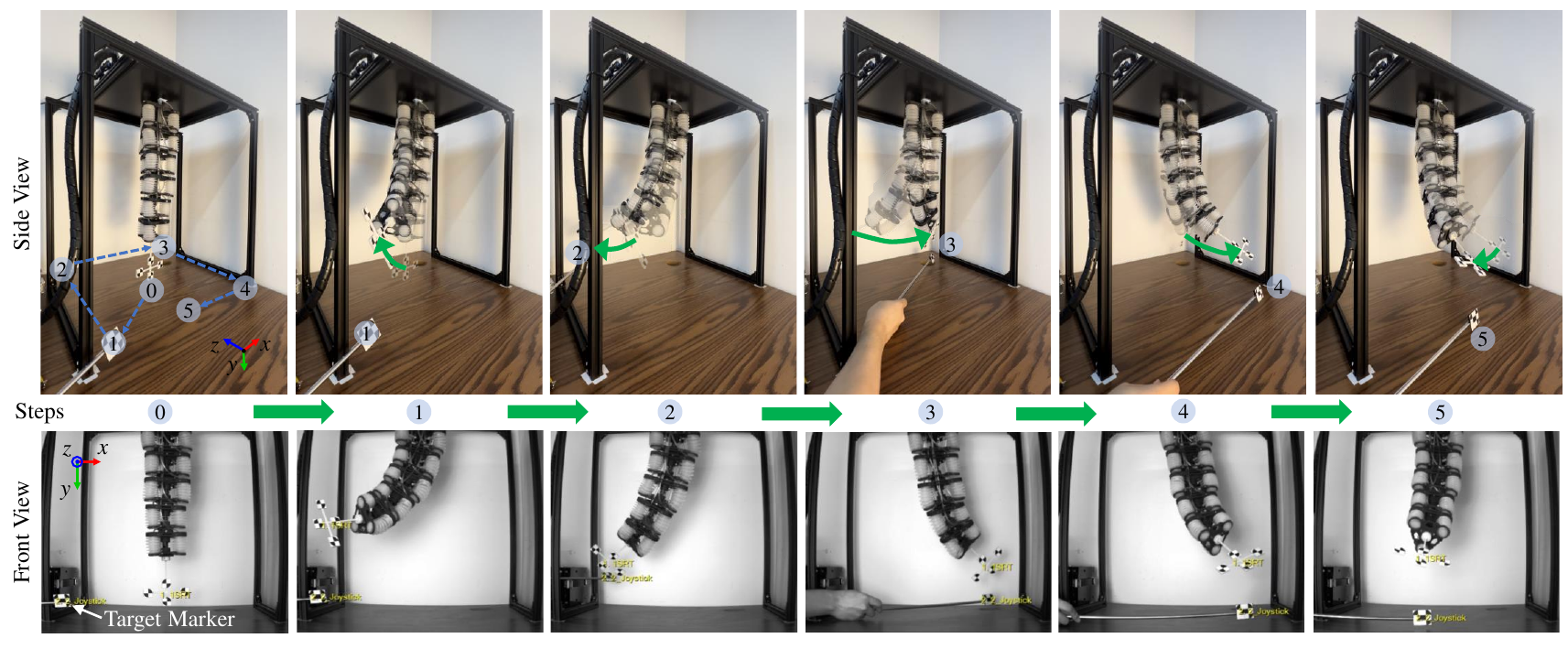}
    \caption{Tracking of moving targets. The soft robot is guided using a marker to reach the position of target at every step. Step $0$ displays the positions and sequence of the five targets which form a square path.  Steps $1\sim5$ show the soft robot's actual target-reaching performance during this task.}
    \label{fig_9}
\end{figure*}

We evaluated the trajectory tracking performance of the soft robot controlled by DK-MPC and compared it with the performance achieved using the same closed-loop MPC controller based on the RBF-based Koopman model (K-MPC). The robot was tasked with following different reference trajectories. 

Fig.~\ref{4} illustrates the process of the marker of the robot's end-effector tracking a circular trajectory "O": the top image shows the actual experimental operation of the robot; while the bottom picture displays the tracking trajectories based on DK-MPC and K-MPC in comparison with the reference trajectory, as well as their respective errors. It can be observed that DK-MPC followed the circular trajectory without bias, with an average error calculated using the method in \cite{bruder2020data} of $2.79 mm$. In contrast, the K-MPC method exhibited significant deviations when tracking the circular trajectory, with an average error of $22.03 mm$. Therefore, DK-MPC achieved higher control accuracy through deeper linearization of the model and enhanced its precision.

We also tested three additional sets of more complex trajectory-tracking scenarios. As shown in Fig.~\ref{5}, these trajectories consist of the letters `T', `H', and `U'. The detailed average errors for all experiments are provided in Table~\ref{tab:methods_comparison}. It can be concluded that DK-MPC achieves more precise global dynamic tracking, which can benefit future control and applications of soft robots.

\subsection{Tracking of Moving Targets}




We further conduct some investigation to demonstrate the application potential of soft robots controlled by DK-MPC. In practical application, the end-effector of the soft robot can be equipped with a gripper for grasping and transportation\cite{li2024disturbance,jiang2021hierarchical,liu2022touchless}. Such tasks require the identification of the targets' position or the teaching of the soft robot to achieve the desired targets. Here, we focus on verifying the DK-MPC's capability for tracking of moving targets.

Fig. \ref{fig_9} illustrates the workflow: at step $0$, five targets are set, forming a square path along the edge of the robot's workspace to showcase the flexibility of the robot.
 
In the experimental procedure, we provide the soft robot with an additional target marker indicating the target it needs to reach. The soft robot then dynamically tracks it. 
The final result is shown in the sequence of side and front views in Fig. \ref{fig_9}, where the green dashed arrows indicate the path of the soft robot moving from the previous target to the current target. This experiment confirms that the soft robot controlled by DK-MPC can dynamically reach the desired target, which lays the foundation for further practical applications.

\section{Conclusion}



This paper proposes a novel multi-segment soft robot control method, DK-MPC, equipped with a deep Koopman operator and an MPC controller. With sampling data, the deep Koopman operator effectively linearizes the high-dimensional, nonlinear dynamics of the soft robot by embedding the system into a higher-dimensional space. This linearization allows for the application of an MPC controller, which optimizes the control inputs over a predefined time horizon to achieve precise trajectory tracking. Through extensive real-world experiments, the DK-MPC method has proven to reduce tracking errors, showcasing its ability to handle the complexities and uncertainties associated with soft robot dynamics. Future work will focus on further enhancing the model's generalizability in multi-segment soft robot control to achieve dexterous operation.


\section{Acknowledgments}
 This work was jointly funded by the National Natural Science Foundation of China(No. 82090053), the National Natural Science Foundation of China(No. 82090050) and Tsinghua University Initiative Scientific Research Program of Precision Medicine (Project No. 2023ZLA001).



\bibliographystyle{IEEEtran}
\bibliography{references.bib}

\begin{thebibliography}{10}
\providecommand{\url}[1]{#1}
\csname url@samestyle\endcsname
\providecommand{\newblock}{\relax}
\providecommand{\bibinfo}[2]{#2}
\providecommand{\BIBentrySTDinterwordspacing}{\spaceskip=0pt\relax}
\providecommand{\BIBentryALTinterwordstretchfactor}{4}
\providecommand{\BIBentryALTinterwordspacing}{\spaceskip=\fontdimen2\font plus
\BIBentryALTinterwordstretchfactor\fontdimen3\font minus \fontdimen4\font\relax}
\providecommand{\BIBforeignlanguage}[2]{{%
\expandafter\ifx\csname l@#1\endcsname\relax
\typeout{** WARNING: IEEEtran.bst: No hyphenation pattern has been}%
\typeout{** loaded for the language `#1'. Using the pattern for}%
\typeout{** the default language instead.}%
\else
\language=\csname l@#1\endcsname
\fi
#2}}
\providecommand{\BIBdecl}{\relax}
\BIBdecl

\bibitem{rus2015design}
D.~Rus and M.~T. Tolley, ``Design, fabrication and control of soft robots,'' \emph{Nature}, vol. 521, no. 7553, pp. 467--475, 2015.

\bibitem{laschi2022embodied}
C.~Laschi, ``Embodied intelligence in soft robotics: Joys and sorrows,'' in \emph{IOP Conference Series: Materials Science and Engineering}, vol. 1261, no.~1.\hskip 1em plus 0.5em minus 0.4em\relax IOP Publishing, 2022, p. 012002.

\bibitem{tang2022learning}
Z.~Tang, P.~Wang, W.~Xin, and C.~Laschi, ``Learning-based approach for a soft assistive robotic arm to achieve simultaneous position and force control,'' \emph{IEEE Robotics and Automation Letters}, vol.~7, no.~3, pp. 8315--8322, 2022.

\bibitem{kwok2022soft}
K.-W. Kwok, H.~Wurdemann, A.~Arezzo, A.~Menciassi, and K.~Althoefer, ``Soft robot-assisted minimally invasive surgery and interventions: Advances and outlook,'' \emph{Proceedings of the IEEE}, vol. 110, no.~7, pp. 871--892, 2022.

\bibitem{xie2023octopus}
Z.~Xie, F.~Yuan, J.~Liu, L.~Tian, B.~Chen, Z.~Fu, S.~Mao, T.~Jin, Y.~Wang, X.~He \emph{et~al.}, ``Octopus-inspired sensorized soft arm for environmental interaction,'' \emph{Science Robotics}, vol.~8, no.~84, p. eadh7852, 2023.

\bibitem{liu2025embodied}
H.~Liu, D.~Guo, and A.~Cangelosi, ``Embodied intelligence: A synergy of morphology, action, perception and learning,'' \emph{ACM Computing Surveys}, 2025.

\bibitem{george2018control}
T.~George~Thuruthel, Y.~Ansari, E.~Falotico, and C.~Laschi, ``Control strategies for soft robotic manipulators: A survey,'' \emph{Soft robotics}, vol.~5, no.~2, pp. 149--163, 2018.

\bibitem{wang2022control}
J.~Wang and A.~Chortos, ``Control strategies for soft robot systems,'' \emph{Advanced Intelligent Systems}, vol.~4, no.~5, p. 2100165, 2022.

\bibitem{haggerty2023control}
D.~A. Haggerty, M.~J. Banks, E.~Kamenar, A.~B. Cao, P.~C. Curtis, I.~Mezi{\'c}, and E.~W. Hawkes, ``Control of soft robots with inertial dynamics,'' \emph{Science robotics}, vol.~8, no.~81, p. eadd6864, 2023.

\bibitem{roesthuis2016steering}
R.~J. Roesthuis and S.~Misra, ``Steering of multisegment continuum manipulators using rigid-link modeling and fbg-based shape sensing,'' \emph{IEEE transactions on robotics}, vol.~32, no.~2, pp. 372--382, 2016.

\bibitem{jones2006kinematics}
B.~A. Jones and I.~D. Walker, ``Kinematics for multisection continuum robots,'' \emph{IEEE Transactions on Robotics}, vol.~22, no.~1, pp. 43--55, 2006.

\bibitem{gao2016mechanical}
A.~Gao, R.~J. Murphy, H.~Liu, I.~I. Iordachita, and M.~Armand, ``Mechanical model of dexterous continuum manipulators with compliant joints and tendon/external force interactions,'' \emph{IEEE/ASME Transactions on Mechatronics}, vol.~22, no.~1, pp. 465--475, 2016.

\bibitem{padmanabhan2015closed}
R.~Padmanabhan, N.~Meskin, and W.~M. Haddad, ``Closed-loop control of anesthesia and mean arterial pressure using reinforcement learning,'' \emph{Biomedical Signal Processing and Control}, vol.~22, pp. 54--64, 2015.

\bibitem{wang2022data}
X.~Wang and N.~Rojas, ``A data-efficient model-based learning framework for the closed-loop control of continuum robots,'' in \emph{2022 IEEE 5th International Conference on Soft Robotics (RoboSoft)}.\hskip 1em plus 0.5em minus 0.4em\relax IEEE, 2022, pp. 247--254.

\bibitem{laschi2023learning}
C.~Laschi, T.~G. Thuruthel, F.~Lida, R.~Merzouki, and E.~Falotico, ``Learning-based control strategies for soft robots: Theory, achievements, and future challenges,'' \emph{IEEE Control Systems Magazine}, vol.~43, no.~3, pp. 100--113, 2023.

\bibitem{li2024disturbance}
X.~Li, Q.~Xiong, D.~Sui, Q.~Zhang, H.~Li, Z.~Wang, T.~Zheng, H.~Wang, J.~Zhao, and Y.~Zhu, ``Disturbance-adaptive tapered soft manipulator with precise motion controller for enhanced task performance,'' \emph{IEEE Transactions on Robotics}, 2024.

\bibitem{george2017learning}
T.~George~Thuruthel, E.~Falotico, M.~Manti, A.~Pratesi, M.~Cianchetti, and C.~Laschi, ``Learning closed loop kinematic controllers for continuum manipulators in unstructured environments,'' \emph{Soft robotics}, vol.~4, no.~3, pp. 285--296, 2017.

\bibitem{wiese2021transfer}
M.~Wiese, G.~Runge-Borchert, B.-H. Cao, and A.~Raatz, ``Transfer learning for accurate modeling and control of soft actuators,'' in \emph{2021 IEEE 4th International Conference on Soft Robotics (RoboSoft)}.\hskip 1em plus 0.5em minus 0.4em\relax IEEE, 2021, pp. 51--57.

\bibitem{coevoet2017software}
E.~Coevoet, T.~Morales-Bieze, F.~Largilliere, Z.~Zhang, M.~Thieffry, M.~Sanz-Lopez, B.~Carrez, D.~Marchal, O.~Goury, J.~Dequidt \emph{et~al.}, ``Software toolkit for modeling, simulation, and control of soft robots,'' \emph{Advanced Robotics}, vol.~31, no.~22, pp. 1208--1224, 2017.

\bibitem{zhou2024cable}
K.~Zhou, B.~Mao, Y.~Zhang, Y.~Chen, Y.~Xiang, Z.~Yu, H.~Hao, W.~Tang, Y.~Li, H.~Liu \emph{et~al.}, ``A cable-actuated soft manipulator for dexterous grasping based on deep reinforcement learning,'' \emph{Advanced Intelligent Systems}, p. 2400112, 2024.

\bibitem{bruder2021koopman}
D.~Bruder, X.~Fu, R.~B. Gillespie, C.~D. Remy, and R.~Vasudevan, ``Koopman-based control of a soft continuum manipulator under variable loading conditions,'' \emph{IEEE Robotics and Automation Letters}, vol.~6, no.~4, pp. 6852--6859, 2021.

\bibitem{bruder2019nonlinear}
D.~Bruder, C.~D. Remy, and R.~Vasudevan, ``Nonlinear system identification of soft robot dynamics using koopman operator theory,'' in \emph{2019 International Conference on Robotics and Automation (ICRA)}.\hskip 1em plus 0.5em minus 0.4em\relax IEEE, 2019, pp. 6244--6250.

\bibitem{wang2022improved}
J.~Wang, B.~Xu, J.~Lai, Y.~Wang, C.~Hu, H.~Li, and A.~Song, ``An improved koopman-mpc framework for data-driven modeling and control of soft actuators,'' \emph{IEEE Robotics and Automation Letters}, vol.~8, no.~2, pp. 616--623, 2022.

\bibitem{bruder2020data}
D.~Bruder, X.~Fu, R.~B. Gillespie, C.~D. Remy, and R.~Vasudevan, ``Data-driven control of soft robots using koopman operator theory,'' \emph{IEEE Transactions on Robotics}, vol.~37, no.~3, pp. 948--961, 2020.

\bibitem{shi2023koopman}
L.~Shi, Z.~Liu, and K.~Karydis, ``Koopman operators for modeling and control of soft robotics,'' \emph{Current Robotics Reports}, vol.~4, no.~2, pp. 23--31, 2023.

\bibitem{chen2023control}
J.~Chen, Y.~Dang, W.~Huo, N.~Yu, and J.~Han, ``Control-oriented modeling of a soft manipulator using the learning-based koopman operator,'' in \emph{2023 29th International Conference on Mechatronics and Machine Vision in Practice (M2VIP)}.\hskip 1em plus 0.5em minus 0.4em\relax IEEE, 2023, pp. 1--6.

\bibitem{shi2022deep}
H.~Shi and M.~Q.-H. Meng, ``Deep koopman operator with control for nonlinear systems,'' \emph{IEEE Robotics and Automation Letters}, vol.~7, no.~3, pp. 7700--7707, 2022.

\bibitem{ng2004feature}
A.~Y. Ng, ``Feature selection, l 1 vs. l 2 regularization, and rotational invariance,'' in \emph{Proceedings of the twenty-first international conference on Machine learning}, 2004, p.~78.

\bibitem{10610952}
Y.~Luo, Q.~Sima, T.~Ji, F.~Sun, H.~Liu, and J.~Zhang, ``Smooth computation without input delay: Robust tube-based model predictive control for robot manipulator planning,'' in \emph{2024 IEEE International Conference on Robotics and Automation (ICRA)}, 2024, pp. 10\,429--10\,435.

\bibitem{luo2024robust}
Y.~Luo, T.~Ji, F.~Sun, Q.~Sima, H.~Liu, M.~Jing, and J.~Zhang, ``Robust tube-based mpc with smooth computation for dexterous robot manipulation,'' \emph{Science China Information Sciences}, vol.~67, no.~11, pp. 1--17, 2024.

\bibitem{goodfellow2016deep}
Y.~LeCun, Y.~Bengio, and G.~Hinton, ``Deep learning,'' \emph{nature}, vol. 521, no. 7553, pp. 436--444, 2015.

\bibitem{jiang2021hierarchical}
H.~Jiang, Z.~Wang, Y.~Jin, X.~Chen, P.~Li, Y.~Gan, S.~Lin, and X.~Chen, ``Hierarchical control of soft manipulators towards unstructured interactions,'' \emph{The International Journal of Robotics Research}, vol.~40, no.~1, pp. 411--434, 2021.

\bibitem{liu2022touchless}
W.~Liu, Y.~Duo, J.~Liu, F.~Yuan, L.~Li, L.~Li, G.~Wang, B.~Chen, S.~Wang, H.~Yang \emph{et~al.}, ``Touchless interactive teaching of soft robots through flexible bimodal sensory interfaces,'' \emph{Nature communications}, vol.~13, no.~1, p. 5030, 2022.

\end{thebibliography}
\end{document}